\documentclass{tlp}
\submitted{February 14, 2016}
\revised{March 21, 2016}
\accepted{March 21, 2016}
\usepackage{bbding}

\usepackage{comment}
\usepackage{wrapfig}
\usepackage{booktabs}
\usepackage[unicode=true, pdfusetitle,
 bookmarks=true,bookmarksnumbered=false,bookmarksopen=false,
 breaklinks=false,pdfborder={0 0 0},backref=false,colorlinks=false]
 {hyperref}
 \let\proof\relax
 
 \usepackage{amsmath}
\usepackage{amsfonts}

\usepackage{xspace}
\newcommand{\mc}[1]{\mathcal{#1}}
\newcommand{\T}{T}
\newcommand{\pS}{\mathcal{S}}

\newcommand{\subtask}[1]{\stepcounter{st}\subsubsection*{Subtask \arabic{st}: #1}}   \newcommand{\subtasksbegin}{}     \newcommand{\subtasksend}{}
\newcommand{\newtext}[1]{{\color{red} #1}}

\renewcommand{\newcommand}{\providecommand}
\usepackage{import}

\usepackage{ifthen}
\usepackage{url}
\usepackage{amssymb}
\usepackage{amsmath}
\usepackage{import}
\usepackage{xparse}
\usepackage{amsmath}
\usepackage[usenames,dvipsnames]{xcolor}
\usepackage{xspace}
\usepackage{datatool}
\usepackage{glossaries}

\providecommand\m[1]{\ensuremath{#1}\xspace}
\renewcommand{\m}[1]{\ensuremath{#1}\xspace}
\newcommand{\trval}[1]{\m{{\bf #1}}}

	\newcommand{\lrule}{\m{\leftarrow}}
	\newcommand{\cause}{\m{\stackrel{c}{\lrule}}}
	
	\newcommand{\ltrue}{\trval{t}}
	\newcommand{\lfalse}{\trval{f}}
	\newcommand{\lunkn}{\trval{u}}
	
	\newcommand{\Tr}{\ltrue}
	\newcommand{\Fa}{\lfalse}
	\newcommand{\Un}{\lunkn}

	\newcommand{\voc}{\m{\Sigma}}

	\newcommand{\struct}{\m{I}}
	
	\newcommand{\I}{\m{\mathcal{I}}}

	\NewDocumentCommand\inter{g+g}{%
	  \IfNoValueTF{#1}
	    {\struct}
	    {\m{#1^{#2}}}}

	\newcommand{\ddd}{\m{\overline{d}}}

	\renewcommand{\int}{\m{\mathbb{Z}}}

	\newcommand{\leqp}{{\m{\,\leq_p\,}}}
	\newcommand{\geqp}{{\m{\,\geq_p\,}}}

	\NewDocumentCommand\subs{g+g}{%
	  \IfNoValueTF{#1}
	    {\m{/}}
	    {\m{#1/ #2}}}

	\newcommand{\logicname}[1]{\textsc{#1}\xspace}

	\newcommand{\idp}{\logicname{IDP}}

	\newcommand{\fodot}{\logicname{FO(\ensuremath{\cdot})}}

\newcommand{\ouracronym}[3]{%
	\newacronym{#1}{#2}{#3}
	\expandafter\newcommand\csname #1\endcsname{\gls{#1}\xspace}%
}
	\ouracronym{FO}{FO}{first-order logic}
	\ouracronym{MX}{MX}{Model Expansion}
	\ouracronym{MO}{MO}{Model Optimization}
	\ouracronym{ASP}{ASP}{Answer Set Programming}
	\ouracronym{TP}{TP}{Theorem Proving}
	\ouracronym{LP}{LP}{Logic Programming}
	\ouracronym{CP}{CP}{Constraint Programming}
	\ouracronym{FP}{FP}{Functional Programming}
	\ouracronym{KR}{KR}{Knowledge Representation}
	\ouracronym{CSP}{CSP}{Constraint Satisfaction Problem}
	\ouracronym{SMT}{SMT}{SAT Modulo Theories}
	\ouracronym{KBS}{KBS}{Knowledge Base System}
	\ouracronym{NNF}{NNF}{Negation Normal Form}
	\ouracronym{FNNF}{FNNF}{Flat Negation Normal Form}
	\ouracronym{DefNNF}{DefNNF}{Definition Negation Normal Form}
	\ouracronym{CDCL}{CDCL}{Conflict-Driven Clause-Learning}
	\ouracronym{LCG}{LCG}{Lazy Clause Generation}
	\ouracronym{AEL}{AEL}{Autoepistemic Logic}
	\ouracronym{OEL}{OEL}{Ordered Epistemic Logic}
	\ouracronym{AFT}{AFT}{Approximation Fixpoint Theory}

	\makeatletter
	\def\ifenv#1{
	\def\@tempa{#1}%
	\def\@ttempa{#1*}%
	\ifx\@tempa\@currenvir
	\expandafter\@firstoftwo
	\else
	\expandafter\@secondoftwo
	\fi
	}
	\makeatother

	\newcommand{\ddrule}[4]{\ensuremath{#1 \leftarrow #2 & \{#3\} & #4}}
	\newcommand{\drule}[2]{\ensuremath{#1 & \leftarrow & #2}}

	\newcommand{\darule}[4]{\ensuremath{#1 \leftarrow #2 & \{#3\} & #4}}
	\newcommand{\arule}[2]{\ensuremath{#1 \, &\leftarrow \, #2}}

	\newcommand{\LNDRule}[2]{
	\ifenv{array}
	{\drule{#1}{#2}}
	{ \ifenv{align}
		{\arule{#1}{#2}}
		{\ifenv{align*}
		{\arule{#1}{#2}}
		{ERROR: using LDRule in unsupported environment: \@currenvir}
		}
	}
	}

	\newcommand{\LDRule}[4]{
	\ifenv{array}
	{\ddrule{#1}{#2}{#3}{#4}}
	{ \ifenv{align}
		{\darule{#1}{#2}{#3}{#4}}
		{\ifenv{align*}
		{\darule{#1}{#2}{#3}{#4}}
		{ERROR: using LDRule in unsupported environment: \@currenvir}
		}
	}
	}

	\NewDocumentCommand\LRule{m+g+g+g}{%
		\IfNoValueTF{#2}%
		{#1.&}{%
		\IfNoValueTF{#3}
		{\LNDRule{#1}{#2.}}
		{\LDRule{#1}{#2.}{#3}{#4}}%
		}
	}

	\NewDocumentCommand\CLRule{m+g}{%
	\ifenv{array}
	{\cdrule{#1}{#2}}
	{ \ifenv{align}
		{\carule{#1}{#2}}
		{\ifenv{align*}
			{\carule{#1}{#2}}
			{ERROR: using CLRule in unsupported environment: \@currenvir}
		}
	}
	}

	\NewDocumentCommand\carule{m+g}{%
		\IfNoValueTF{#2}
			{\ensuremath{#1.}}
			{\ensuremath{#1 \, &\cause \, #2}}}
	\NewDocumentCommand\cdrule{m+g}{%
		\IfNoValueTF{#2}
			{\ensuremath{#1.}}
			{\ensuremath{#1 & \cause & #2}}}

	\newcommand{\algrule}[4]{
	\hbox{{#1}:}& 
	\quad #2 ~\longrightarrow~ #3 
	\hbox{~ if } #4\\
	}

	\newcommand{\AlgoRule}[4]{
	\ifenv{array}
	{\algrule{#1}{#2}{#3}{#4}}
		{ERROR: using AlgoRule in unsupported environment: \@currenvir}
	}

\newcommand{\commentstyle}{\color{Gray}}

	\usepackage[final]{listings}

	\lstdefinelanguage{idp}{
		morekeywords=[1]{query(}, 
		morekeywords=[2]{namespace,vocabulary,theory,structure,procedure,term,set,formula, spec, specification,query},
		morekeywords=[3]{include,using,type,isa,contains,partial,extern,LFD,GFD,constructed,from,constraint,pred,supertype,of,subtype,define},
		morekeywords=[4]{int,float,char,string,nat},
		morekeywords=[5]{if,then,else,for,end},
		morecomment=[s]{/*}{*/},	
		morecomment=[l]{//}
	}
	\newcommand{\ignore}[1]{}

	\newboolean{nocomments}
	\setboolean{nocomments}{false}

	\newboolean{commentmargin}
	\setboolean{commentmargin}{true}

	\newcommand{\namedcomment}[3]{
		\ifthenelse{\boolean{nocomments}}
		{} 
		{ 
			\ifthenelse{\boolean{commentmargin}}
				{ {\color{#3} \marginpar{\color{#3}\sc #2}#1}  } 
				{  {\color{#3} {\sc #2}: #1}  } 
		}
	}
	\newcommand{\mnamedcomment}[3]{\ifthenelse{\boolean{nocomments}}{}{{\marginpar{ \color{#3}{\sc #2}:#1}}}}

	\newcommand{\todo}[1]{\namedcomment{#1}{TODO}{blue}}

	\usepackage{soul}

\usepackage{etoolbox}

\newcommand\setcitation[2]{%
  \csdef{mycommoncitation#1}{#2}}
\newcommand\getcitation[1]{%
  \csuse{mycommoncitation#1}}

\setcitation{IDP}{WarrenBook/DeCatBBD14}
\setcitation{idp}{WarrenBook/DeCatBBD14}
\setcitation{fodot}{tocl/DeneckerT08}
\setcitation{CP}{Apt03}
\setcitation{EZCSP}{lpnmr/Balduccini11}
\setcitation{KR}{Baral:2003}
\setcitation{ASPComp2}{lpnmr/DeneckerVBGT09}
\setcitation{ASPComp3}{journals/tplp/CalimeriIR14}
\setcitation{ASPComp4}{conf/lpnmr/AlvianoCCDDIKKOPPRRSSSWX13}
\setcitation{CPSupport}{ictai/DeCat13}
\setcitation{CPsupport}{ictai/DeCat13}
\setcitation{functionDetection}{iclp/DeCatB13}
\setcitation{fodot2asp}{DeneckerLTV12} 
\setcitation{Tarskian}{DeneckerLTV12} 
\setcitation{Inca}{iclp/DrescherW12}
\setcitation{csp2asp}{ijcai/DrescherW11}
\setcitation{DPLLT}{cav/GanzingerHNOT04}
\setcitation{AspInPractice}{synthesis/2012Gebser}
\setcitation{clasp}{ai/GebserKS12}
\setcitation{oclingo}{kr/GebserGKOSS12}
\setcitation{gringo}{lpnmr/GebserST07}
\setcitation{cmodels}{aaai/GiunchigliaLM04}
\setcitation{inputster}{tplp/Jansen13}
\setcitation{DLV}{tocl/LeonePFEGPS06}
\setcitation{LearningPaper}{TPLP/BruynoogheBBDDJLRDV} 
\setcitation{clog}{iclp/BogaertsVDV14} 
\setcitation{foc}{iclp/BogaertsVDV14}
\setcitation{inferenceClog}{ecai/BogaertsVDV14}
\setcitation{examplesClog}{nmr/BogaertsVDV14b} 
\setcitation{AFT}{DeneckerBV12}
\setcitation{KBS}{iclp/DeneckerV08}
\setcitation{KBPE}{inap/DePooterWD11}
\setcitation{lazyGrounding}{jair/CatDBS15} 
\setcitation{lazygrounding}{jair/CatDBS15} 
\setcitation{ASP}{marek99stable}
\setcitation{satid}{sat/MarienWDB08}
\setcitation{lazyclausegeneration}{constraints/OhrimenkoSC09}
\setcitation{FP}{ACMCS/Hudak89}
\setcitation{GroundingWithBounds}{jair/WittocxMD10}
\setcitation{SAT}{faia/SilvaLM09}
\setcitation{LTC}{iclp/Bogaerts14}
\setcitation{SPSAT}{ictai/DevriendtBMDD12}
\setcitation{BreakID}{cspsat/DevriendtBB14}
\setcitation{LCG}{stuckeyLCG}
\setcitation{MiniZinc}{conf/cp/NethercoteSBBDT07}
\setcitation{minizinc}{conf/cp/NethercoteSBBDT07}
\setcitation{amadini}{cpaior/AmadiniGM13}
\setcitation{bootstrapping}{lash/BogaertsJDJBD14}
\setcitation{GroundedFixpoints}{ai/BogaertsVD15}
\setcitation{LogicBlox}{datalog/GreenAK12}
\setcitation{proB}{journals/sttt/LeuschelB08}
\setcitation{NaturalInductions}{KR/DeneckerV14} 
\setcitation{LP}{jacm/EmdenK76}
\setcitation{SMT}{faia/BarrettSST09}
\setcitation{AF}{ai/Dung95}
\setcitation{ADF}{kr/BrewkaW10}
\setcitation{ADFRevisited}{ijcai/BrewkaSEWW13}
\setcitation{DefaultLogic}{ai/Reiter80}
\setcitation{DL}{ai/Reiter80}
\setcitation{AEL}{mo85}
\setcitation{minisat}{sat/EenS03}
\setcitation{completion}{adbt/Clark78}
\setcitation{wasp}{lpnmr/AlvianoDFLR13}
\setcitation{minisatid}{ictai/DeCat13}
\setcitation{lcg}{stuckeyLCG}
\setcitation{CEGAR}{jacm/ClarkeGJLV03}
\setcitation{cegar}{jacm/ClarkeGJLV03}
\setcitation{CuttingPlane}{or/DantzigFJ54}
\setcitation{kodkod}{tacas/TorlakJ07}
\setcitation{cdcl}{Marques-SilvaS99}
\setcitation{CDCL}{Marques-SilvaS99}
\setcitation{}{}
\setcitation{}{}
\setcitation{}{}
\setcitation{}{}
\setcitation{}{}
\setcitation{}{}
\setcitation{}{}
\setcitation{}{}
\setcitation{}{}
\setcitation{}{}
\setcitation{}{}
\setcitation{}{}
\setcitation{}{}
\setcitation{}{}
\setcitation{}{}
\setcitation{}{}
\setcitation{}{}
\setcitation{}{}
\setcitation{}{}
\setcitation{}{}
\setcitation{}{}
\setcitation{}{}
\setcitation{}{}
\setcitation{}{}
\setcitation{}{}
\setcitation{}{}
\setcitation{}{}
\setcitation{}{}
\setcitation{}{}
\setcitation{}{}
\setcitation{}{}
\setcitation{}{}
\setcitation{}{}
\setcitation{}{}
\setcitation{}{}
\setcitation{}{}
\setcitation{}{}
\setcitation{}{}
\setcitation{}{}
  
\newcommand\mycite[1]{%
      \ifcsname mycommoncitation#1\endcsname%
   \cite{\getcitation{#1}}%
  \else%
    \cite{#1}
  \fi%
}	
  
\usepackage{amsthm}

\theoremstyle{plain}
\newtheorem{thm}{Theorem}[section]

\newtheorem*{lem*}{Lemma}

\theoremstyle{definition}

\newtheorem{definition}[thm]{Definition}

\newtheorem*{nota*}{Notation}

\newtheorem*{note*}{Note}

\newtheoremstyle{example_basic} 
{\topsep} 
{\topsep} 
{\normalfont}
{0pt}
{\bfseries}
{.}
{5pt plus 1pt minus 1pt}
{}

\newtheoremstyle{example_contd}
{\topsep} 
{\topsep} 
{\normalfont}
{0pt}
{\bfseries}
{.}
{5pt plus 1pt minus 1pt}
{\thmname{#1} \thmnumber{ #2}\thmnote{#3} (continued)}

\theoremstyle{example_basic} 

\newtheorem{example}[thm]{Example}

\newtheorem{ex*}{Example}
\theoremstyle{example_contd}

\theoremstyle{plain}

 \renewcommand{\S}{S}
\renewcommand{\I}{I}
\newcommand{\citet[1]}{\citeN{#1}\xspace}
\setboolean{nocomments}{true}\renewcommand{\newtext}[1]{#1}
\newcommand\blfootnote[1]{  \begingroup
  \renewcommand\thefootnote{}\footnote{#1}  \addtocounter{footnote}{-1}  \endgroup
}
\begin{document}
\title[The KB paradigm in interactive configuration]{The KB paradigm and its application to interactive configuration.}
\author[P. Van Hertum et al.]
{Pieter Van Hertum, Ingmar Dasseville, Gerda Janssens, Marc Denecker \\
Department of Computer Science\\
KU LEUVEN\\
Leuven, BELGIUM\\
first.lastname@cs.kuleuven.be
}

\maketitle              \blfootnote{This is an extended version of a paper presented at the international
symposium on Practical Aspects of Declarative Languages (PADL 2016),
invited as a rapid communication in TPLP. The authors acknowledge the
assistance of the conference program chairs Marco Gavanelli and John Reppy. This research was supported by the project GOA 13/010 Research Fund KU Leuven and projects G.0489.10, G.0357.12,
and G.0922.13 of the Research Foundation - Flanders.}

\begin{abstract}

The knowledge base paradigm aims to express domain knowledge in a rich formal language, and to use this domain knowledge as a knowledge base to solve various problems and tasks that arise in the domain by applying multiple forms of inference.
As such, the paradigm applies a strict separation of concerns between information and problem solving.
In this paper, we analyze the principles and feasibility of the knowledge base paradigm in the context of an important class of applications: interactive configuration problems.
In interactive configuration problems, a configuration of interrelated objects under constraints is searched, where the system assists the user in reaching an intended configuration.
It is widely recognized in industry that good software solutions for these problems are very difficult to develop.
We investigate such problems from the perspective of the KB paradigm. 
We show that multiple functionalities in this domain can be achieved by applying different forms of logical inferences on a formal specification of the configuration domain.
We report on a proof of concept of this approach in a real-life application with a banking company.
To appear in Theory and Practice of Logic Programming (TPLP).

\end{abstract}

\section{Introduction}
In this paper, we investigate the application of knowledge representation and reasoning (KRR) to the problem of {\em interactive configuration}. 
In the past decades enormous progress in many different areas of computational logic was obtained. 
This resulted in a complex landscape with many declarative paradigms, languages and communities. 
One issue that fragments computational logic more than anything else is the reasoning/inference task. 
Computational logic is divided in different declarative paradigms, each with its own syntactical style, terminology and conceptuology, and designated form of inference (e.g, deductive logic, logic programming, abductive logic programming, databases (query inference), answer set programming (answer set generation), constraint programming, etc.). 
Yet, in all of them declarative propositions need to be expressed. 
Take, e.g., ``each lecture takes place at some time slot''. 
This proposition could be an expression to be deduced from a formal specification if the task was a verification problem, or to be queried in a database, or it could be a constraint for a scheduling problem. 
It is, in the first place, just a piece of information and we see no reason why depending on the task to be solved, it should be expressed in a different formalism (classical logic, SQL, ASP, MiniZinc, etc.).

The Knowledge Base (KB) paradigm \mycite{KBS} was proposed as an answer to this. 
The KB paradigm applies a strict separation of concerns to information and problem solving.
A KB system allows information to be stored in a knowledge base, and provides a range of inference methods.
With these inference methods various types of problems and tasks can be solved using the {\em same knowledge base}. 
As such the knowledge base is neither a program nor a description of a problem, it cannot be executed or run. 
It is nothing but information. However, this information can be used to solve multiple sorts of problems. 
Many declarative problem solving paradigms are mono-inferential: they are based on one form of inference. 
In comparison, the KB-paradigm is multi-inferential. 
We believe that this implements a more natural, pure view of what declarative logic is aimed to be.
The $\fodot$  KB project \mycite{KBS} is a research project that runs now for a number of years.
Its aim is to integrate different useful language constructs and forms of inference from different declarative paradigms in one rich declarative language and a KB system.  
So far, it has led to the KB language $\fodot$~\mycite{fodot}  and the KB system \idp~\mycite{IDP} which were used in the configuration experiment described in this paper.

An interactive configuration (IC) problem \cite{ai/McDermott82,ijcai/MittalF89,FleischanderlFHSS98,aiedam/JunkerM03,cp/Hadzic04} is an interactive version of a constraint solving problem. 
One or more users search for a configuration of objects and relations between them that satisfies a set of constraints.
Industry abounds with interactive configuration problems: configuring composite physical systems such as cars and computers, insurances, loans, schedules involving human interaction, webshops (where clients choose composite objects), etc. 
However, building such software is renowned in industry as difficult and no broadly accepted solution methods are available~\cite{Felfernig14,jlp/AxlingH96}. 
Building software support using standard imperative programming is often a nightmare~\cite{cacm/BarkerO89,hicss/PillerHIS14}, due to the fact that (1) many functionalities need to be provided, (2) they are complex to implement, and (3) constraints on the configuration tend to get duplicated and spread out over the application, in the form of snippets of code performing various computations relative to the constraint (e.g., context dependent checks or propagations) which often leads to an unacceptable maintenance cost. 
This makes interactive configuration an excellent domain to illustrate the advantages of declarative methods over standard imperative or object-oriented  programming.

Our research question is: can we express the constraints of correct
configurations in a declarative logic and provide the required
functionalities by applying inference on this domain knowledge?  This
is a KRR question albeit a difficult one. In the first place, some of the
domain knowledge may be complex. For an example in the context of a
computer configuration problem, take the following constraint: {\em
the total memory usage of different software processes that needs to be in main
memory simultaneously, may not exceed the available RAM memory}. It
takes an expressive knowledge representation language with aggregates to (compactly
and naturally) express such a constraint. Many interactive
configuration problems include complex constraints: various sorts
of quantification, aggregates, definitions
(sometimes inductive), etc. Moreover, an interactive configuration
system needs to provide many functionalities: checking the validity of
a fully specified configuration, correct and safe reasoning on a
partially specified configuration (this involves reasoning on
incomplete knowledge, sometimes with infinite or unknown domains),
computing impossible values or forced values for attributes,
generating sensible questions to the user, providing explanation why certain values are
impossible, backtracking if the user regrets some choices, supporting
the user by filling in his don't-cares while potentially taking into account a
cost function, etc.  

That declarative methods are particularly suitable for solving this type of problem has been acknowledged before, and several systems and languages have been developed~\cite{cp/Hadzic04,inap/SchneeweissH11,TiihonenHAS13,ppdp/VlaeminckVD09}. 
A first contribution of this paper
is the analysis of IC problems from a Knowledge Representation point of view.
We show that multiple functionalities in this domain can be achieved by applying different forms of logical inference on {\em the same}  formal specification of the configuration domain. We define various sorts of inference and analyse them in terms of which different functionalities can be supplied.
The second contribution is the reverse: we  study the feasibility and usefulness of the KB paradigm in this important class of applications.
The logic used in this experiment is
the logic \fodot~\mycite{fodot}, an extension of first-order logic
(FO), and the system is the IDP system~\mycite{IDP}.  We discuss the
complexity of (the decision problems of) the inference problems and
why they are solvable, despite the high expressivity of the language
and the complexity of inference. 
This research has its origin in an experimental IC system we developed in collaboration with industry.
We evaluated our approach using the evaluation criteria of the knowledge-based configuration research~\cite{Felfernig14}.
We conclude this paper with a discussion of related work in using knowledge-based systems for configuration and a comparison of our approach with these systems. \section{The FO(.) KB project}

\label{ssec:fodot}
\label{sec:kbs}

\paragraph{The language.} 
$\fodot$  refers to the class of extensions of first order logic (FO) as is common in logic, e.g. FO(LFP) stands for the extension of FO with a least fixpoint construction \cite{cav/ImmermanV97}.
Currently, the language of the \idp system in the project is FO(T, ID, Agg, arit, PF) \cite{tocl/DeneckerT08,tplp/PelovDB07}: FO extended with types, definitions, aggregates, arithmetic and partial functions.
Abusing notation, we will use $\fodot$ as an abbreviation for this language. Below, we introduce the aspects of the logic and its syntax on which this paper relies.

\paragraph{A specification.} A \textit{vocabulary} is a set $\Sigma$ of type (denoted as $\Sigma_T$), predicate (denoted as $\Sigma_P$) and function symbols (denoted as $\Sigma_F$). 
Variables $x, y$, atoms $A$, FO-formulas $\varphi$ are defined as usual. 
A predicate $P$ of arity $n$ has a type $[\tau_1,\ldots,\tau_n]$, a $n$-tuple of type symbols.
A function of arity $n$ has a type $[\tau_1,\ldots,\tau_n]\rightarrow \tau_{n+1}$, a $(n+1)$-tuple of type symbols.
Aggregate terms are of the form $Agg(E)$, with $Agg$ an aggregate function symbol and $E$ an expression $\{(\overline{x},F(\overline{x}))|\varphi(\overline{x})\}$, where $\varphi$ is any FO-formula, $F$ a function symbol and
$\overline{x}$ a tuple of variables. Examples are the cardinality,
sum, product, maximum and minimum aggregate functions. For example $sum\{(x,F(x))|\varphi(x)\}$ is read as $\Sigma_{x\in\{y|\varphi(y)\}} F(x)$. A \textit{term}
in $\fodot$ can be an aggregate term or a term as defined in FO. 
A \textit{theory} is a set of $\fodot$ formulas.

A \emph{partial set} on domain $D$ is a function from $D$ to $\{\Tr,\Un,\Fa\}$. 
A partial set is two-valued (or total) if $\Un$ does not belong to its range. 
A \textit{(partial) structure} $\pS$ consists of a domain $D_\tau$ for all types $\tau$ in $\Sigma_T$ and an assignment of a partial set $\sigma^\pS$ to each predicate or function symbol $\sigma\in(\Sigma_P\cup\Sigma_F)$, called the \emph{interpretation} of $\sigma$ in $\pS$.  
The interpretation $P^\pS$ of a predicate symbol $P$ with type $[\tau_1,\ldots,\tau_n]$ in $\pS$ is a
partial set on domain $D_{\tau_1}\times \ldots \times D_{\tau_n}$. For
a function $F$ with type $[\tau_1, \ldots, \tau_n]\rightarrow
\tau_{n+1}$, the interpretation $F^\pS$ of $F$ in $\pS$ is a partial
set on domain $D_{\tau_1}\times \ldots \times D_{\tau_n} \times
D_{\tau_{n+1}}$. In case the interpretation of (a predicate or function symbol) $\sigma$ in $\pS$ is a
two-valued set, we abuse notation and use $\sigma^\pS$ as shorthand for
$\{\ddd|\sigma^\pS(\ddd)=\Tr\}$. 
The precision-order on the truth values is given by $\Un<_p \Fa$ and $\Un<_p \Tr$. 
It can be extended  pointwise to partial sets  and partial structures, denoted $\pS\leq_p\pS'$. 
Informally, this means that an interpretation has become more precise if tuples of domain elements that were previously mapped to unknown now map to true or false.
Notice that total structures are the maximally precise ones. We will illustrate this precision relation in Example \ref{ex:specification}. 
We say that $\pS'$ extends $\pS$ if $\pS\leq_p \pS'$. 
We will sometimes use $\sigma^\pS_{x}$ as shorthand for the set $\{\ddd|\ddd \in D_{\tau_1}\times \ldots \times D_{\tau_n} \land \sigma^\pS(\ddd)=x\}$, with $x\in \{\Tr,\Fa,\Un\}$.

\newtext{
The associated theory $T_\pS$ of a partial structure $\pS$ is a representation of the information contained in $\pS$ as a theory, which will be used in Section \ref{sec:solution}. 
It is defined by the following collection of constraints. 
For every predicate symbol $P$, this collection contains two sets of constraints:
\begin{align*}
&\{P(\overline{d})|\ddd \in P^\pS_\Tr\}\\
&\{\neg P(\overline{d})|\ddd \in P^\pS_\Fa\}
\end{align*}
and two sets of constraints for every function symbol $F$:
\begin{align*}
&\{F(\overline{d})=e|(\ddd,e) \in F^\pS_\Tr\}\\
&\{\neg F(\overline{d})=e|(\ddd,e) \in F^\pS_\Fa\}
\end{align*}
Given a partial structure $\pS$, the domain structure $\pS_D$ is the structure containing only the domains of $\pS$. It is easy to see that $\pS$ contains the same information as $\T_\pS \cup \pS_D$.
}
A total structure\footnote{Note the difference in typography between a partial structure $\pS$ and a total structure $\S$.} $\S$  is called {\em functionally consistent} if for each function $F$ with type $[\tau_1, \ldots, \tau_n]\rightarrow
\tau_{n+1}$, the interpretation $F^\S$ is the graph of a function $D_{\tau_1} \times \ldots \times D_{\tau_n} \mapsto D_{\tau_{n+1}}$. 
A partial structure $\pS$ is functionally consistent if it has a functionally consistent two-valued extension. 
Unless stated otherwise, we will assume for the rest of this paper that all (partial) structures are functionally consistent.

A domain atom (domain term) is a tuple of a predicate symbol $P$ (a function symbol $F$) and a tuple of domain elements $(d_1,\ldots,d_n)$. 
We will denote it as $P(d_1,\ldots,d_n)$ (respectively $F(d_1,\ldots,d_n)$).
We say a domain term $t$ of type $\tau$ is uninterpreted in $\pS$ 
if  $\{d|d \in D_{\tau} \land (t=d)^\pS= \Un\}$ is non-empty.

To define the satisfaction relation on theories, we extend the interpretation of symbols to arbitrary terms and formulas using the Kleene truth assignments~\cite{Kleene52}. 
For a theory $\T$ and a partial structure $\pS$, we say that $\pS$ is a model of $\T$ (or in symbols $\pS\vDash \T$) if $\T^\pS=\Tr$ and $\pS$ is two-valued.
We sometimes abuse notation and write $\T\vDash \varphi$ for the entailment relation, as a shorthand for ``For every structure $\S$ such that $\S\vDash \T$, we have $\S\vDash \varphi$.''.

\begin{example}\label{ex:specification}
 To illustrate some of the concepts introduced above, assume a situation where we have some knowledge about printers, that have some type of connection.
 A vocabulary to model such knowledge can look as follows:
    \begin{align*}
     &\voc =\{ \\
     & \quad \Sigma_T=\{printer, connection\}\\
     & \quad \Sigma_P=\{PrinterConnection(printer,connection)\}\\
     & \quad \Sigma_F=\{ \}\\ \}
    \end{align*}
A structure $\pS_0$ in which we have 2 printers $P_1$ and $P_2$ and 2 possible connections: $USB$ and $LAN$, where we have no additional information, looks like:
    \begin{align*}
     &\pS_0=\{ \\
     & \quad printer = \{P_1,P_2\}\\
     & \quad connection = \{USB, LAN\}\\
     & \quad PrinterConnection = \{ 
(P_1,USB)\rightarrow \Un, (P_2,USB)\rightarrow \Un,\\
& \qquad (P_1,LAN)\rightarrow \Un, (P_2,LAN)\rightarrow \Un\}
    \\ \}
    \end{align*}
A more precise structure $\pS_1\geq_p \pS_0$, containing the partial information that $P_1$ has $USB$ and $P_2$ certainly has no $LAN$ connection looks like:
    \begin{align*}
     &\pS_1=\{ \\
     & \quad printer = \{P_1,P_2\}\\
     & \quad connection = \{USB, LAN\}\\
     & \quad PrinterConnection = \{ 
(P_1,USB)\rightarrow \Tr, (P_2,USB)\rightarrow \Un,\\
& \qquad (P_1,LAN)\rightarrow \Un, (P_2,LAN)\rightarrow \Fa\}
    \\ \}
    \end{align*}
A total structure $\S_2 \geq_p \pS_1$ containing full information can look like:
    \begin{align*}
     &\S_2=\{ \\
     & \quad printer = \{P_1,P_2\}\\
     & \quad connection = \{USB, LAN\}\\
     & \quad PrinterConnection = \{ 
(P_1,USB)\rightarrow \Tr, (P_2,USB)\rightarrow \Tr,\\
& \qquad (P_1,LAN)\rightarrow \Fa, (P_2,LAN)\rightarrow \Fa\}
    \\ \}
    \end{align*}
\end{example}

\paragraph{Inference tasks.} \label{ssec:inferences} 
In the KB paradigm, a specification is a bag of information. This information can be  used for solving various  problems by applying a suitable form of inference on it. 

FO is standardly associated with deduction inference: a deductive
inference task takes as input a pair of theory $\T$ and sentence
$\varphi$, and returns $\Tr$ if $\T\models\varphi$ and $\Fa$
otherwise. This is well-known to be undecidable for FO, and by extension
for $\fodot$. However, to provide the required functionality of an
interactive configuration system we can use simpler forms of
inference. Indeed, in many such domains a fixed finite domain is
associated with each unknown configuration parameter.  

A natural format in logic to describe these finite domains is by a
partial structure with a finite domain. Also other data that are often
available in such problems can be represented in that structure. As
such various inference tasks are solvable by finite domain reasoning
and become decidable. Below, we give the base forms of inference for our KB system and
recall their complexity when using finite domain reasoning.  We assume a fixed  vocabulary $\Sigma$ and theory $\T$ and query. Our complexities are given in function of the domain size.
\begin{description}
   \item[Modelexpand($\T,\pS$):] input:  theory ${\T}$ and partial structure $\pS$;
     output: a model $\I$ of $\T$ such that $\pS\leqp\I$ or \textit{UNSAT} if there is no such $\I$. 
     Modelexpand~\cite{lash08/WittocxMD08} is a generalization  for $\fodot$ theories of the modelexpansion task as defined in Mitchell et al.~\cite{MitchellT05}.
     Complexity  of deciding the existence of a modelexpansion is in \textbf{NP}. 
     Structure $\S_2$ in Example \ref{ex:specification} is the output of Modelexpand($\T,\pS_1$), with $\pS_1$ as in Example \ref{ex:specification}, and $\T$ a theory consisting of the constraint that every printer has exactly one connection.
 \item[Modelcheck($\T,\S$):] input: a total structure $\S$ and theory $\T$ over the vocabulary interpreted by $\S$; output is the boolean value $\S \models \T$. 
 Note that Modelcheck is a degenerate case of the Modelexpand inference, with $\pS$ a total structure.  Complexity is in \textbf{P}.

   \item[Minimize($\T,\pS,t$):] input: a theory $\T$, a partial structure
     $\pS$ and a term $t$ of numerical type; output: a model $\I\geqp \pS$
     of $\T$ such that the value $t^{\I}$ of $t$ is minimal. The term $t$ represents a numerical expression whose value has to be minimized.
     This is an extension to the modelexpand inference.
     The complexity of deciding that      a certain $t^\I$ is minimal, is in $\mathbf{\Delta_2^P}$.
   \item[Propagate($\T,\pS$):]  input:  theory ${\T}$ and partial structure $\pS$; output: the most precise partial structure $\pS_r$ such that for every model $\I\geqp\pS$ of ${\T}$ it is true that $I\geq_p \pS_r$.
   The complexity of deciding that a partial structure $\mc{S}'$ is  $\mc{S}_r$ is in $\mathbf{\Delta_2^P}$. Note that we assume that all partial structures are functionally consistent, which implies that we also propagate functional integrity constraints.

\item[Query($\pS,E$):] input: a (partial) structure $\pS$ and a set expression
  $E=\{\overline{x}\mid \varphi(\overline{x})\}$;   output: the set $A_Q=\{\overline{x}\mid\varphi(\overline x)^\pS=\Tr\}$. Complexity of deciding that a set $A$ is $A_Q$ is in
    \textbf{P}.

\end{description}

Approximative versions exist for some of these inferences, with lower complexity \cite{ppdp/VlaeminckVD09}.
More inferences exist, such as simulation of temporal theories in
\fodot~\mycite{LTC}, but were not
used in the experiment.

\section{Interactive Configuration}
In an IC problem, one or more users search for a configuration of objects and relations between them that satisfies a set of constraints. 

Typically, the user is not aware of all constraints. 
There may be too many of them to keep track of.  
Even if the human user can oversee all constraints that he needs to satisfy, he is not a perfect reasoner and cannot comprehend all consequences of his choices.  
This in its own right makes such problems hard to solve.  
The problems get worse if the user does not know about the relevant objects and relations or the constraints on them, or if the class of involved objects and relations is large, if the constraints get more complex and more ``irregular" (e.g., exceptions), if more users are involved, etc.  
On top of that, the underlying constraints in such problems tend to evolve quickly.
All these complexities occur frequently, making the problem difficult for a human user. 
In such cases, computer assistance is needed: the human user chooses and the system assists by guiding him through the search space.  

For a given IC problem, an IC system has information on that problem. 
There are a number of stringent rules to which a configuration should conform, and besides this there is a set of parameters.
Parameters are the open fields in the configuration that need to be filled in by the user or decided by the system.
\subsection{Running example: Domain knowledge}
A simplified version of the application in Section \ref{implementation} is used in Section \ref{sec:solution} as running example. 
We introduce the domain knowledge of this example here.
\noindent \begin{example}
\label{MotEx}
Software on a computer has to be configured for different employees.
Table \ref{tab:ex} contains the information on the software, the requirements, the budgets of the employees and the prices of software.
Available software is Windows, Linux, \LaTeX, Office and a DualBoot system.
Each software item has a price, which can be seen in column \textbf{PriceOf}.
Column \textbf{PreReq} specifies which software is required for other software. 
Every type of employee has a budget, provided in column \textbf{MaxCost}.
\textbf{IsOs} lists the pieces of software that are operating systems.
Next to the information in the table, we know that if more than one OS is installed, a DualBoot System is required.
\begin{table}
\caption{Example data \label{tab:ex}}
\centering
\begin{tabular}{llcllcllcl}
\toprule 
\multicolumn{2}{c}{\textbf{PriceOf}} 	& \hspace{0.4cm} 	& \multicolumn{2}{c}{\textbf{PreReq}}	 	& \hspace{0.4cm} & \multicolumn{2}{c}{\textbf{MaxCost}} 	&\hspace{0.4cm} 	&\multicolumn{1}{c}{\textbf{IsOS}}\\
\textit{software} & \textit{int} 	& \hspace{0.4cm} 	& \textit{software} & \textit{software}		& \hspace{0.4cm} & \textit{employee}	 & \textit{int} 	&\hspace{0.4cm} 	&\textit{software}\\
\midrule 
Windows 	& 60 		& &  Office	& Windows	& & Secretary 	& 100 		& & Windows	\\
Linux 		& 20 		& &  \LaTeX	& Linux		& & Manager 	& 150 		& & Linux	\\
\LaTeX 		& 10		& &  		&  		& & 		&     		& & 		\\
Office		& 30		& &  		&  		& & 		&     		& & 		\\
DualBoot 	& 40		& &  		&  		& & 		&     		& & 		\\
\bottomrule
\end{tabular}
\end{table}

\end{example}
\newcounter{st}[section]
\subsection{Subtasks of an interactive configuration system}\label{ssec:subtasks}
Any system assisting a user in interactive configuration must be able to perform a set of subtasks. 
We look at important subtasks that an interactive configuration system should support.
\subtasksbegin
\subtask{Acquiring information from the user}
The first task of an IC system is acquiring information from the user.
The system needs to get a value for a number of parameters of the configuration from the user.
There are several options: the system can ask questions to the user, it can make the user fill in a form containing open text fields, dropdown-menus, checkboxes, etc.
Desirable aspects would be to give the user the possibility to choose the order in which he gives values for parameters and to omit filling in certain parameters (because he does not know or does not care).
For example, in the running example a user might need a \LaTeX-package, but he does not care about which OS he uses. 
In that case the system will decide in his place that a Linux system is required.
Since a user is not fully aware of all constraints, it is possible that he inputs conflicting information.
This needs to be handled or avoided.

\subtask{Generating consistent values for a parameter}
After a parameter is selected (by the user or the system) for which a value is needed, the system can assist the user in choosing these values.
A possibility is that the system presents the user with a list of all possible values, given the values for other parameters and the constraints of the configuration problem.
Limiting the user with this list makes that the user is unable to input inconsistent information. 

\subtask{Propagation of information}
Assisting the user in choosing values for the parameters, a system can use the constraints to propagate the information that the user has communicated.
This can be used in several ways. A system can communicate propagations through a GUI, for example by coloring certain fields red or graying out certain checkboxes.
Another way is to give a user the possibility to explicitly ask ``what if''-questions to the system.
In Example \ref{MotEx}, a user can ask the system what the consequences are if he was a secretary choosing an Office installation.
The system answers that in this case a Windows installation is required, which results in a Linux installation becoming impossible (due to budget constraints) and as a consequence it  also derives the impossibility of installing \LaTeX.

\subtask{Checking the consistency for a value}
When it is not possible/desirable to provide a list of possible values, the system checks that the value the user has provided is consistent with the known data and the constraints.

\subtask{Checking a configuration}
If a user makes manual changes to a configuration, the system provides him with the ability to check if his updated version of the configuration still conforms to all constraints.

\subtask{Autocompletion}
If a user has finished communicating all his preferences, the system autocompletes the partial configuration to a full configuration.
This can be done arbitrarily (a value for each parameter such that the constraints are satisfied) or the user can have some other parameters like total cost, that have to be optimized.

\subtask{Explanation}
If a supplied value for a parameter is not consistent with other parameters, the system can explain this inconsistency to the user.
This can be done by showing minimal sets of parameters with their values that are inconsistent, by showing (visualizations of) constraints that are violated or by combinations of both.
It can also explain to the user why certain automatic choices are made, or why certain choices are impossible.

\subtask{Backtracking}
It is not unthinkable that a user makes a mistake, or changes his mind after seeing consequences of choices he made. 
Backtracking is an important subtask for a configuration system, and can be supported in numerous ways.
The simplest way is a simple back button, which reverts the last choice a user made.
A more involved option is a system where a user can select any parameter and erase his value for that parameter.
The user can then decide this parameter at a later timepoint.
Even more complex is a system where a user can supply a value for a parameter and if it is not consistent with other parameters the system shows him which parameters are in conflict and proposes other values for these parameters such that consistency can be maintained.

\subtasksend

\section{Interactive Configuration in the KB paradigm}\label{sec:solution}
 To analyze the IC problem from the KB point of view, we aim at formalizing the subtasks of Section 3 as inferences.
 In this paper we do not deal with user interface aspects.
 For a given application, our knowledge base consists of a vocabulary $\voc$, a theory $\T$ expressing the configuration constraints and a partial structure $\pS$.
 Initially, $\pS_0$ is the partial structure that contains the domains of the types and the input data.
 During IC, $\pS_0$ will become more and more precise partial structures $\pS_i$ due to choices made by the user. 
 For IC, the KB also contains $L_{\pS_0}$, the set of all uninterpreted domain atoms/terms\footnote{In the rest of this paper, a domain atom is treated as a term that evaluates to true or false.}  in $\pS_0$.
 These domain terms are the logical formalization of the parameters of the IC problem.
 $\voc$ and $\T$ are fixed.
 As will be shown in this section, all subtasks can be formalized by (a combination of) inferences on this knowledge base consisting of $\Sigma, \T, \pS_0, L_{\pS_0}$ and information gathered from the user. 
\begin{example}\label{ExD}
    \newtext{Continuing Example \ref{MotEx}, use vocabulary $\voc$:
    \begin{align*}
     &\voc = \\
     & \quad \Sigma_T=\{software,\ employee, \ int\}\\
     & \quad \Sigma_P=\{ Install(software),\ IsOS(software),\ PreReq(software,software)\}\\
     & \quad \Sigma_F=\{PriceOf(software):int,\ MaxCost(employee):int,\\
     & \quad \quad \quad \quad \quad Cost:int,\ Requester:employee\}
    \end{align*}
    The initial partial structure $\pS_0$ consists of:
    \begin{align*}
      &employee \rightarrow \{Secretary,  Manager\}\\
      &software \rightarrow \{Windows, Linux, LaTeX, Office, DualBoot\}
    \end{align*}
  and interpretations for \textit{MaxCost (employee):int, IsOs(software), PriceOf(software): int} and \textit{PreReq(software, software)} as can be seen in Table \ref{tab:ex}. 
  All symbols from $\Sigma$ that are not specified above are assumed to be fully unknown in $\pS_0$.\\
    The set of parameters $L_{\pS_0}$ is: 
    \begin{align*}
\{&Requester, Install(Windows), Install(Linux),\\ & Install(Office), Install(LaTeX), Install (DualBoot), Cost\}     
    \end{align*}
    The theory $\T$ consists of the following constraints:
    \[\begin{array}{l}  
      \forall s1\, s2:\   Install(s1)\land PreReq(s1,s2)\Rightarrow Install(s2).\\
     \quad \text{// The total cost is the sum of the prices of all installed software.}\\ 
			Cost = sum\{(s,PriceOf(s)) | Install(s)\}.\\
      Cost \leq MaxCost(Requester).\\
      \exists s:\ Install(s)\land IsOS(s).  \\
      Install(Windows)\land Install(Linux)  \Rightarrow Install(DualBoot).
      \end{array}\]}
 \end{example}
\subtasksbegin
\subtask{Acquiring information from the user}
Key in IC is collecting information from the user on the parameters.
During the run of the system, the set of parameters that are still open changes. 
In our KB system a derived inference (a combination of the inferences as introduced in Section \ref{ssec:inferences}) is used to calculate this set of parameters.
Complexity results of derived inferences stem from basic results formulated by \citet{MitchellT05} and the observation that modelchecking is polynomial in the size of the domain.

 \begin{definition}
   \textbf{Calculating uninterpreted terms.}\\ \textbf{GetOpenTerms($\T,\pS$)} is the derived inference with input a theory $\T$, a partial structure $\pS\geq_p\pS_0$ and the set $L_{\pS_0}$ of terms. 
   Output is a set of terms such that for every term $t$ in that set, there exist models $\I_1$ and $\I_2$ of $\T$ that extend $\pS$ ($\I_1,\I_2\geq_p\pS$) for which $t^{\I_1}\neq t^{\I_2}$. 
   Or formally:
   $$\{l|l\in L_{\pS_0} \land \{d|(l=d)^{\pS'}=\Un\}\neq \emptyset \land \pS'=Propagate(\T,\pS)\}$$
 \end{definition}
  The complexity of deciding whether a given set of terms $A$ is the set of uninterpreted terms is in $\mathbf{\Delta_2^P}$.
 
An IC system can use this set of terms  in a number of ways. It can use a metric to select a specific term, which it can pose as a direct question to the user.
It can also present a whole list of these terms at once and let the user pick one to supply a value for.
In Section \ref{implementation}, we discuss two different approaches we implemented for this project.

 \begin{example}
    \label{ex:questions}  
    In Example \ref{ExD}, the parameters and domains are already given.
    Assume that the user has chosen the value \textit{Manager} for \textit{Requester}, true for \textit{Install(Windows)} and false for \textit{Install(Linux)}. 
    The system will return \textit{GetOpenTerms}$(\T,\pS)$ = $\{$\textit{Install(Office), Install(DualBoot), Cost}$\}$.
  \end{example}
  \subtask{Generating consistent values for a parameter}
A domain element $d$ is a possible value for term $t$ if there is a model $\I\geq_p\pS$ such that $(t=d)^\I=\Tr$.
\begin{definition}
    \textbf{Calculating consistent values.}\\
    \textbf{GetConsistentValues($\T,\pS,t$)} is the derived inference with input a theory $\T$, a partial structure $\pS$ and a term $t\in GetOpenTerms(\T,\pS)$.
    Output is the set $$\{t^{\I}|\   \I \text{ is a model of }\T\text{ extending }\pS  \}$$
  \end{definition}
  The complexity of deciding that a set $P$ is the set of consistent values for $t$ is in $\mathbf{\Delta^P_2}$.
  \begin{example}
    \label{ex:values}  
    The consistent values for $Requester$ given $\T$ and the initial partial structure $\pS_0$ from Example \ref{ExD} is:
    \begin{align*}
     &GetConsistentValues(\T,\pS, Requester)=\{Secretary, Manager\}\\
    \end{align*}
  Consistent values for other terms are the integers for \textit{Cost} and $\{$\textit{true, false}$\}$ for the others. 
  \end{example}
\subtask{Propagation of information}
It is informative for the user that he can see the consequences of assigning a particular value to a parameter. 
\begin{definition}
  \textbf{Calculating Consequences.}\\
  \textbf{PosConsequences($\T,\pS,t,a$)} and \textbf{NegConsequences($\T,\pS,t,a$)} are derived inferences with input a theory $\T$, a partial structure $\pS$, an uninterpreted term $t$ $\in$ GetOpenTerms$(\T,\pS)$ and a domain element $a$ $\in$ GetConsistentValues$(\T,\pS,t)$. As output it has a set $C^+$, respectively $C^-$ of tuples $(q,b)$ of uninterpreted terms and domain elements.
  $(q,b)\in C^+$, respectively $C^-$ means that the choice $a$ for $t$ entails that $q$ will be forced, respectively prohibited to be $b$.
  Formally, 
  \begin{align*}
    C^+=\{(q,b)\ |\ & (q=b)^{\pS'}=\Tr \land (q=b)^{\pS}=\Un  \\
			\land \ & \pS'=Propagate(\T,\pS\cup \{t=a\})\\
			\land \ & q \in GetOpenTerms(\T,\pS) \setminus \{t\} \ \}\\
    C^-=\{(q,c)\ |\ & (q=c)^{\pS'}=\Fa \land (q=c)^{\pS}=\Un  \\
			\land \ & \pS'=Propagate(\T,\pS\cup \{t=a\})\\
			\land \ & q \in GetOpenTerms(\T,\pS) \setminus \{t\} \ \}\
  \end{align*}
\end{definition}
The complexity of deciding whether a set $P$ is $C^+$ or $C^-$ is in $\mathbf{\Delta_2^P}$.
\begin{example}\label{ExAssisting}
  Say the user has chosen $Requester= Secretary$ and wants to know the consequences of making $Install(Windows)$ true.
  The output in this case contains
  $(Install(LaTeX),\Fa)$  in $PosConsequences(\T,\pS,t,a)$ and $(Install(LaTeX),\Tr)$ in  $NegConsequences(\T,\pS,t,a)$ 
since this combination is too expensive for a secretary. 
Note that there is not always such a correspondence between the positive and negative consequences. For example, when deriving a negative consequence for $Cost$, this does not necessarily imply a positive consequence.
\end{example}
    \subtask{Checking the consistency for a value}
A value $d$ for a term $t$ is consistent if there exists a model of $\T$ in which $t=d$ that extends the partial structure representing the current state.
\begin{definition}
    \textbf{Consistency Checking.}\\ \textbf{CheckConsistency($\T,\pS,t,d$)} is the derived inference with input a theory $\T$, a partial structure $\pS$, an uninterpreted term $t$ and a domain element $d$.
    Output is a boolean $b$ that represents whether $\pS$ extended with $t=d$ still satisfies $\T$.
    Formally we return $\Tr$ if
    $$(\pS\cup \{t^\pS=d\})\vDash \T$$
    and $\Fa$ otherwise. Complexity of deciding if a value $d$ is consistent for a term $t$ is in \textbf{NP}.
  \end{definition}
\begin{example}
 If a user has chosen $Install(Windows)$ and $Install(LaTeX)$ to be true, then $Manager$ will and $Secretary$ will not be a consistent answer for $Requester$.
\end{example}

\subtask{Checking a configuration}
 Once the user has constructed a 2-valued structure $\S$ and makes manual changes to it, he may need to check if all constraints are still satisfied. A theory $T$ is checked on a total structure $\S$ by calling $Modelcheck(T,S)$, with complexity in  \textbf{P}.
\subtask{Autocompletion} 
If a user is ready communicating his preferences (Subtask 1) and there are undecided terms left which he does not know or care about, the user may want to get a full configuration (i.e. a total structure). This is computed by modelexpand. In particular: $$\I=Modelexpand(\T,\pS)$$

    In many of those situations the user wants to have a total structure with, for example, a minimal cost (given some term representing the cost $t$). This is computed by minimize:
 $$\I=Minimize(\T,\pS,t)$$
    
   \begin{example}\label{ExCompleting}
      Assume the user is a secretary and all he knows is that he needs Office. 
      He chooses $Secretary$ for $Requester$  and true for $Install(Office)$ and calls autocompletion.
      A possible output is a structure $\S$ where for the remaining parameters, a
      choice is made that satisfies all constraints, e.g.,
      $Install(Windows)^\S=\Tr$, $Install(DualBoot)^\S=\Tr$ and the other $Install$ atoms false.  This is not a cheapest solution (lowest cost). By calling minimize using cost-term $Cost$, the DualBoot is dropped. 
    \end{example} 
    
  \subtask{Explanation}
\newtext{  It is clear that a whole variety of options can be developed to provide different kinds of explanations to a user.
 If a user supplies an inconsistent value for a parameter, options can range from calculating an inconsistent subset of the theory $\T$ (1) to giving a proof of inconsistency as in \cite{iclp/PontelliS06} (2), to calculating a partial subconfiguration that has this inconsistency (3). 
UnsatSubstructure is a logical inference for option 3.
    \begin{definition}\label{def:inconstructure}
      \textbf{Calculating inconsistent structures.}\\
       \textbf{UnsatSubstructure($\T,\pS$)} is a derived inference with input a theory $\T$ and a partial structure $\pS$ that cannot be extended to a model of $\T$ and as output all (partial) structures $\pS'\leq_p \pS$ such that $\pS'$ cannot be extended to a model $\I$ of $\T$. Formally, we return:
      $$\{\pS'|\pS' \leq_p \pS \land \neg (\exists \I \geq_p \pS' \land \I\vDash \T)\}$$
  Complexity of deciding if a set is an inconsistent substructure is in $\mathbf{co-NP}$.
 \end{definition}
 The inference in Definition \ref{def:incontheoryzb} calculates an inconsistent subtheory. 
 \begin{definition}\label{def:incontheoryzb}
     \textbf{Calculating inconsistent theories.}\\
      \textbf{UnsatSubtheory($\T,\pS$)} is a derived inference with input theory $\T$ and a partial structure $\pS$ such that there does not exist a model $I$, extending $S$, satisfying $\T$. The inference has as output all theories $\T'$ such that $\T'\subseteq \T$ and there is no model satisfying $T$, extending $\pS$.
      Formally, we return:
      $$\{\T'|\T'\subseteq \T \land \neg (\exists \I\geq_p \pS \land \I\vDash \T')\}$$
      Complexity of deciding if a theory is such an inconsistent theory is in $\mathbf{co-NP}$.
   \end{definition}
 Note that Definition \ref{def:inconstructure} and \ref{def:incontheoryzb} do not make any statements of minimality. 
 
 Using the associated theory $\T_\pS$ and domains structure $\pS_D$ of a partial  structure $\pS$, it is possible to consider calculating minimally precise partial configurations as a special case of calculating a minimal inconsistent subset of the theory.
 As in \cite{esws/ShchekotykhinFRF14}, we can introduce a ``background theory'' $B\subset \T\cup \T_\S$ (a subset of the theory in which there are assumed to be no conflicts).
 We define multiple derived logical inferences, with different degrees of minimality (not-minimal, subset-minimal and minimal in size) of increasing complexity, able to provide explanations to the user.
     \begin{definition}\label{def:incontheory}
     \textbf{Calculating inconsistent theories with a background.}\\
      \textbf{UnsatSubtheory($\T,\pS,B$)} is a derived inference with input theory $\T$, a partial structure $\pS$ and a background theory $B\subseteq \T \cup T_\pS$ such that there does not exist a model $I$, with the domains as in $\pS_D$ satisfying $\T\cup\T_\pS$ (or equivalently: extending $\pS$ and satisfying $\T$), but there is a model satisfying $B$. The inference has as output all theories $\T'$ such that $B\subseteq \T'\subseteq \T\cup\T_\pS$ and there is no model satisfying $T$.
      Formally, we return:
      $$\{\T'|B\subseteq \T'\subseteq (\T\cup\T_S) \land \neg (\exists \I\geq_p \pS_D \land \I\vDash \T')\}$$
      Complexity of deciding if a theory is such an inconsistent theory is in $\mathbf{co-NP}$.
   \end{definition}
   \begin{definition}\label{def:sminincontheory}
    \textbf{Calculating minimal inconsistent theories with a background.}\\
      \textbf{MinimalUnsatTheory($\T,\pS,B$)} is a derived inference with input theory $\T$, a partial structure $\pS$ and a background theory $B$ as above.
      Output is the subset of subset minimal theories from \textit{UnsatSubtheory($\T,\pS,B$)}.
			                  Complexity of deciding if a set is a subset minimal inconsistent theory is in $\Delta^P_2$.
   \end{definition}
   \begin{definition}\label{def:cminincontheory}
    \textbf{Calculating minimum inconsistent theories with a background.}\\
      \textbf{MinimumUnsatTheory($\T,\pS,B$)} is a derived inference with input theory $\T$, a partial structure $\pS$ and a background theory $B$ as above.
      Output is the subset of cardinality minimal theories from \textit{MinimalUnsatTheory($\T,\pS,B$)}.
      Complexity of deciding if a set is a cardinality minimal inconsistent theory is $\Pi^P_2$.
   \end{definition} 

    Note that Definition \ref{def:inconstructure} is equivalent to calculating a minimal inconsistent subset of a theory $\T\cup \T_\S$, with $B=\T$, if you translate the output back to a pair of a theory and a structure. 
    Definition \ref{def:incontheoryzb} is equivalent to calculating a minimal inconsistent subset of a theory $\T\cup \T_\pS$, with $B=\T_\pS$, if you translate the output back to a pair of a theory and a structure. 

 In recent literature multiple approaches are discussed, all mapping to one of our explanation-related inferences. QuickXPlain \cite{aaai/Junker04} is an algorithm that implements Definition \ref{def:incontheory}. The Hitting Set Directed Acyclic Graph (HSDAG) \cite{ai/Reiter87} algorithm calculates subset minimal inconsistent theories (Definition \ref{def:sminincontheory}, as in different ASP solvers \cite{kbse/ShlyakhterSJST03,nmr/Syrjanen06}.  Implementations of Definition \ref{def:cminincontheory} have been described in \cite{conf/sat/LynceM04} and \cite{conf/ausai/ZhangLS06}. In our experiment, we have an implementation of Definition \ref{def:sminincontheory} \cite{iclp/WittocxVD09}, where we do however do not calculate the entire set of subset minimal theories. We only calculate one, which gives one explanation of the inconsistency. If the user resolves that problem, he can ask for a new explanation which will point to another reason of inconsistency. This process is reiterated untill all problems are resolved.
 
 \begin{example}
  We show a minimal inconsistent subtheory in a situation with $\T$ as in Example $\ref{ExD}$ and $\pS_i$, a partial structure representing an intermediate configuration where a user started with $\pS_0$ and has chosen $Secretary$ for $Requester$, and wants to Install \textit{Office} and \textit{Linux}. This is not possible, and as such, the user asks the system for an explanation in the form of a minimal inconsistent theory. A possible minimal inconsistent theory with $B=\emptyset$, is: 
  \begin{align*}
     & (Install(\text{\textit{Office}}) \land PreReq(\text{\textit{Office}},Windows)) \Rightarrow Install(Windows).\\
     &Cost = sum\{(s,PriceOf(s)) | Install(s)\}.\\
     &Cost  \leq  MaxCost(Requester).
  \end{align*}
This means that there is no valid configuration because Windows needs to be installed as prerequisite for Office, and the total cost then exceeds the budget of a Secretary.

 \end{example}

}
  \subtask{Backtracking}
If a value for a parameter is not consistent, the user has to choose a new value for this parameter, or backtrack to revise a value for another parameter.
In Section \ref{ssec:subtasks} we discussed three options of increasing complexity for implementing backtracking functionality.
Erasing a value for a parameter is easy to provide in our KB system, and since this is a generalization of a back button (erasing the last value) we have a formalization of the first two options.
Erasing a value $d$ for parameter $t$ in a partial structure $\pS$ is simply modifying $\pS$ such that $(t=d)^\pS=\Un$.
As with explanation, a number of more complex options can be developed. We look at one possibility.
Given a partial configuration $\pS$, a parameter $p$ and a value $d$ that is inconsistent for that parameter, calculate a minimal set of previous choices that need to be undone such that this value is possible for this parameter. The converse of this problem is very well known under the name of maximum satisfiability problems. In other words, you want to hold on to as much of the structure as possible while ensuring satisfiability. 

This problem is closely related to the explanation subtask \cite{conf/aaai/HerasMM11,conf/date/Marques-SilvaP08}. You can imagine the explanation problem as asking the system to point out a mistake in your reasoning. However, solving this mistake will not guarantee you have not made any other mistake in the rest of the problem. What we actually need is a minimal set of things we can remove, so every problem is solved simultaneously. 

So more formally, we can use Definition \ref{def:inconstructure} and calculate $UnsatStructure(\T\land (t=d), \pS)$. This inference calculates a set $A$ of sets of previous choices that together are inconsistent. 
Undoing an arbitrary choice in all of these sets results in a partial subconfiguration $\pS'$ of $\pS$ such that $d$ is a possible value for $t$ in $\pS'$. To find the maximal partial subconfiguration $\pS'$ that satisfies that property, the minimal hitting set~\cite{ai/Reiter87} of all sets in $A$ has to be calculated.

\subtasksend
 
\section{Proof of Concept}
  
\subsection{Implementation}\label{implementation}
In this section we will describe the developed application and implementation.
Our application has a simple client-server architecture. The server plays the role of the reasoning engine, which is mainly a thin wrapper around the \idp system. The client consists of a GUI made in QML~\cite{url:qml} as front-end.

The server converts \idp into a stateless server which is accessible through the web. The client application sends the necessary information, consisting of theories, partial structures and choices, to this server and the server executes the needed inferences. This is a design which involves repeatedly sending over the choices a user has made, but it allows for a very simple architecture to show the feasibility of our design.

This implementation was done in cooperation with Adaptive Planet, a consulting company~\cite{url:adaptiveplanet}
that developed the user interface, and an international banking company that provided us with a substantial configuration problem for testing purposes. 
More practical information about this implementation, some screenshots, a 
    downloadable demo and another example of a configuration system developed with \idp as a reasoning engine (a simpler course configuration demo) can be found at: \url{http://www.configuration.tk}.

    \subsubsection{The Reasoning Engine}

As explained before, the application we developed was built on the knowledge base system \idp, which was not developed specifically with configuration problems in mind.
It provides the basic inferences listed at the end of Section \ref{sec:kbs}. 
The goal of this experiment was to check if this general infrastructure could be readily applied to applications such as configuration. 

In Section \ref{sec:solution} we showed how the tasks which are needed for configuration relate to the infrastructure provided by \idp. 
Our main implementation task was to convert these specifications to code. 
Some subtasks such as autocompletion did not require any extra work, as this functionality is directly available as the modelexpand inference. 
Some functionality, e.g. calculating consequences, did require some work but the existing functionality provided almost all needed components. 

We mainly use the existing forms of inference that are readily available in the \idp system. No dedicated or specialized algorithms are used for the configuration subtasks. This proves the point that the KB-paradigm is very flexible but this also means that we had relatively little impact upon the efficiency of our server. 
However, the system ended up being quite responsive and we could conclude that \idp (and by extension the KB-paradigm) passed the test for usefulness in this application.

\subsubsection{User Interface}
    Apart from a reasoning engine, it is also necessary to have an accessible front end so the user has easy access to the multitude of functionalities which are available.
The front end consists of an application written in the Qt framework using QML~\cite{url:qml} and connects to a configuration engine over the web. 
For the purposes of our demo, we developed two different graphical interfaces: 

  \paragraph{Wizard}
  In the wizard interface, the user is interrogated and he answers on subsequent questions selected by the system, using the $GetOpenTerms$ inference. 
  An important side note here is that the user can choose not to answer a specific question, for instance because he cannot decide as he is missing relevant information or because he is not interested in the actual value (at this point). 
  These parameters can be filled in at a later timepoint by the user, or by the system, using propagation, or in case the user calls autocompletion.

  \paragraph{Drill-Down}
  In the drill-down interface, the user sees a list of the still open parameters, and can pick which one he wants to fill in next. 
  This interface is useful if the user is a bit more knowledgeable about the specific configuration and wants to give the values in a specific order.
  
  \vspace{1em}
  In both interfaces the user is assisted in the same way when he enters data. 
  When he or the system selects a parameter, he is provided with a dropdown list of the possible values, using the $GetConsistentValues$ inference.
  Before committing to a choice, he is presented with the consequences of his choice, using the calculate consequences inference.
  The nature of the system guarantees a correct configuration and will automatically give the user support using all information it has (from the knowledge base, or received from the user).  
  \subsection{Evaluation}\label{sec:evaluation}
\subsubsection{Evaluation Criteria}  \label{ssec:evaluationcriteria}
When evaluating the quality of software (especially when evaluating declarative methods), scalability (data complexity) is often seen as the most important quality metric.
Naturally when using an interactive configuration system, performance is important.
However, in the configuration community it is known that reasoning about typical configuration problems is relatively easy and does not exhibit real exponential behavior \cite{TiihonenHAS13}.
Also, depending on the application, it is reasonable to expect the number of parameters to be limited, since humans need to fill in the configuration in the end.
When developing a configuration system, challenges lie in the complexity of the knowledge, its high volatility and the complex functionalities to be built. 
To get a more complete view of the performance of a configuration system, we chose to evaluate on a larger set of different evaluation criteria.
\newtext{In recent literature \cite{Felfernig14} nine evaluation criteria are used to differentiate between different paradigms used for configuration.
In Section \ref{sec:related}, ten other approaches will be discussed and compared to our solution using the same nine criteria.
   \begin{description}
\item[Grapical Modeling Concepts (C1)] is supported if there are standard graphical modeling techniques available that visualize configuration knowledge. They improve understandability, development time and maintenance of new knowledge bases.
\item[Component Oriented modeling (C2)] is a criterion that states that the modeling language is a natural language that allows knowledge base design on the basis of real-world concepts: types, relations, hierarchies, etc.
\item[Automated Consistency Maintenance (C3)]  can be broken down to two categories.
Firstly, a system can have support for a priori automated consistency maintenance. 
This helps a developer write consistent constraints and verifying correctness while writing the knowledge base.
Secondly, runtime automated consistency maintenance supports the end user, by guaranteeing that every intermediate configuration he can make, can be extended to a valid configuration.
\item[Modularization concepts are available (C4)] if the modeling language is modular and has support for adding additional structure to the knowledge base, for example by organizing the constraints in blocks or groups.
\item[Maintainability (C5)] relates to the adaptability of the knowledge base if the background information changes.
This background information is volatile, it is for example depending on ever-changing company policies.
As such, it is vital that when that information changes, the system can be easily adapted. 
When using custom software, all tasks using domain knowledge (like rules and policies) need their own program code.
The domain knowledge is scattered all over the program. 
If this policy changes, a programmer has to find all snippets of program code that are relevant for guarding this policy and modify them. 
This results in a system that is hard to maintain, hard to adapt and error-prone.
Every time the domain knowledge changes, a whole development cycle has to be run through again.
Some systems have support for intelligent knowledge base navigation tools for complex knowledge spaces.
\item[Model-based (C6)] means that a knowledge base in the system expresses exactly what it means for a configuration to be valid. This in contrast to rule-based configuration, where a knowledge base also contains problem solving knowledge (i.e. information on how the rules should be used/fired).
\item[Efficiency (C7)] relates to efficiency and scalability of the reasoning engine. 
\item[Ability to solve generative problem settings (C8)] means that the language supports talking about component types instead of specific objects. 
A system supports generic constraints if it allows for constraints that apply to every instance of a component type on which the constraint is defined.
For example, the first constraint of Theory $T$ in Example \ref{ExD} is a generic constraint about all software, without explicitly naming the individual pieces of software.
\item[Ability to provide explanations (C9)] means that the system is able to communicate reasons for inconsistencies or explain why certain choices are forced/prohibited.
\end{description}
}

\subsubsection{Evaluation}\label{ssec:evaluation}
\newtext{
The criteria discussed in previous section are a good way to evaluate the KB implementation of a configuration system. 
We evaluate our implementation and the \idp system with these criteria.
\begin{description}
\item[Grapical Modeling Concepts (C1).] 
\idp has no support for graphical modeling of domain knowledge and we did not develop any tools for this experiment.
However, it must be noted, that a highly expressive and readable modeling language often makes graphical modeling obsolete.
\item[Component Oriented modeling (C2).] The \fodot language used in this experiment is an extension of typed first-order logic. 
First-order logic is about a small set of connectives: $\land, \lor, \neg, \Rightarrow, \Leftrightarrow, \exists, \forall$. These connectives are also the basic connectives of information used by humans. 
Classical logic is a good KR language because it has a very clear informal semantics. 
It does however not suffice for knowledge representation. 
\fodot extends classical logic with a number of extensions that arise from research in AI and KR, such as aggregates, inductive definitions, types, \ldots
This makes \fodot a suited modeling language for a configuration system.
\item[Automated Consistency Maintenance (C3).] 
A priori consistency maintenance is supported in the implementation by using the explanation inferences.
If the developer has a collection of constraints that is consistent, it is possible to evaluate if a new constraint leads to an inconsistency and ask the system what other constraints it conflicts with, using for example definition
\ref{def:sminincontheory}. 
At runtime consistency maintenance is partially supported, by using the inferences in subtask 2, 3 and 4.
These inferences are theoretically able to guarantee consistency, but due to computational limitations, approximate versions can be used. These are not always able to give the same guarantees.
\item[Modularization concepts are available (C4). ]
The implemented configuration system is modular, since a knowledge base can consist of multiple theories and structures, that together make up the specification. 
The explanation inference allows that a user selects  background constraints, as in definition \ref{def:sminincontheory}, and in this way he can choose about which constraints he needs feedback. 
\item[Maintainability (C5).] 
The development of a KB system with a centrally maintained knowledge base makes the knowledge directly available, readable and adaptable. 
A well-known advantage of this approach is in maintainability: if domain information changes, the developer can easily modify the knowledge base.
The current implementation does however have no additional support for knowledge base navigation tools.
\item[Model-based (C6).]
The \fodot modeling methodology is based on formulating the properties of a correct configuration in a natural way, such that the models of a specification correspond with configurations. 
This is inherently a model-based approach.
\item[Efficiency (C7).] As explained in Section \ref{implementation}, we have only written a thin layer upon existing software which did not target configuration problems specifically. 
The performance of the \idp system has been tested extensively in other contexts \cite{ppdp/JansenDDJ14,TPLP/BruynoogheBBDDJLRDV}. 
The reasoning engine for \idp is very similar in performance to mainstream ASP solvers~\cite{journals/tplp/CalimeriIR14}. Their performance was tested more extensively in the context of configuration by \citet{TiihonenHAS13}. 
It is also very difficult to reliably compare the response times for interactive systems. 
Standard benchmarking techniques in software engineering traditionally use instances which need multiple minutes to solve. 
In this setting we aim for subsecond response times, for which no standard benchmarks are available as far as we are aware. 

In this experiment (a configuration task with 300 parameters and 650 constraints), our users reported a response time of a half second on average with outliers up to 2 seconds.
Note that the provided implementation was a naive prototype and optimizing the efficiency of the implemented algorithms is still possible in a number of ways.

\item[Ability to solve generative problem settings (C8).] 
$\fodot$ is an extension of first-order logic, and as such has native support for quantification which is needed for generative problem settings.
\item[Ability to provide explanations (C9).] Subtask 7 and 8 in Section  \ref{sec:solution} are inferences that are used to support giving explanations.
The implemented configuration system has an implementation of definition \ref{def:sminincontheory}.
\end{description}
}

\todo{PIETER: ASP resultaten vermelden}

 \section{Related Work}

\label{sec:related} 
\subsection{Other approaches}
In different branches of AI research, people have been focusing on configuration software in different settings. 
The following discussion of knowledge-based approaches is based on a book in recent literature~\cite{Felfernig14}.
After the discussion we will compare the ten approaches with our approach (\textbf{IDP}). 

Historically, the first knowledge-based configuration systems were \emph{rule-based} (\textbf{RBS}) \cite{ai/McDermott82,cacm/BarkerO89}. 
These systems operate on a working memory and if the condition of a rule is fulfilled, it fires and modifies the working memory, applying the conclusion of that rule.
Rule-based systems are sensitive to rule orderings. This complicates modification of the rule-base.
More importantly, inclusion of problem solving knowledge in the rule-base, makes a rule-base problem specific and focused towards one specific task.
This leads to the same problems as in imperative languages. To solve different tasks, more rule-bases have to be built, leading to duplication and fanning out of knowledge, giving issues in maintainability.

\emph{Constraint Satisfaction Problems}  are widely used for tackling configuration problems~\cite{ijcai/MittalF89,FleischanderlFHSS98}. 
A (static\footnote{In constrast to dynamic and generative constraint satisfaction problem.}) constraint satisfaction problem (\textbf{SCSP}) is a triple $(V,D,C)$ of a set of domain variables $V=\{v_1, v_2, \ldots, v_n\}$, a set of domains $\{dom(v_1), dom(v_2), \ldots, dom(v_n)\}$ and set of constraints $C$.
A solution for a SCSP is an assignment $S$ of domain elements $d_i\in dom(v_i)$ to variables $v_i$, such that each variable has a value in $S$ and constraints $C$ are satisfied by $S$.
A configuration task in SCSP is searching for a solution for a SCSP $(V,D,C)$, where $C$ contains the configuration constraints together with the user preferences.
To make efficient CSP configuration systems, different techniques have been used, such as local search \cite{Li2005}, symmetry breaking \cite{ki/KiziltanFH01} and knowledge compilation techniques such as binary decision diagrams \cite{cp/HadzicA05}. 
In response to limitations of SCSP in configuration, extensions have been developed.
\emph{Dynamic Constraint Satisfaction Problems} (\textbf{DCSP})~\cite{aaai/MittalF90} allow for variables to be inactive or irrelevant. 
If a variable is inactive, it does not need a value in a solution (for example, when configuring a smartphone, no camera resolution is needed if no camera is present).
\emph{Generative Constraint Satisfaction Problems} (\textbf{GCSP})~\cite{FleischanderlFHSS98} extends SCSP with component types and generative constraints. 

\citet{splc/Janota08}  studied a mapping of CSP to SAT to use a SAT solver to provide functionality for a configuration system.

There exist many graphical approaches for doing knowledge configuration, and visualizing a configuration model.
\citet{kang1990feature} used \emph{feature models} (\textbf{FM}) for modeling these concepts, while \textbf{UML} was proposed in \cite{aicom/FalknerH13}.
FM and UML configuration approaches have no reasoning algorithms, they need to be used with external algorithms.
\citet{splc/KaratasOD10} for example combined feature models with constraint logic programming (CLP) to provide reasoning and automated analysis.

Decidable subsets of first-order logic, \emph{description logics} (\textbf{DL}) are used often in context of the semantic web. 
They have also been used for the development of configuration systems \cite{HotzKDSN06,McGuinnessW98}.
The trade-off for having decidable subsets of first-order logic is that they are limited in expressivity. This make domain knowledge in these systems less readable, less natural and harder to maintain.
An ontology based method was also proposed by Vanden Bossche et al. \citeyear{VandenBosscheRMVP07} using OWL. 

Tiihonen et al. developed a configuration system WeCoTin~\cite{TiihonenHAS13}, based on \emph{Answer Set Programming} (\textbf{ASP}).
WeCoTin uses Smodels, an ASP system, as inference engine, for propagating consequences of choices. 
Answer set programming (ASP) is a form of declarative programming based on the stable-model semantics~\cite{iclp/GelfondL88} for logic programs. 
The architecture of their reasoning engine is closely related to the reasoning engine we use. 
Also, in language, many similarities can be identified~\cite{DeneckerLTV12}, as they both have their roots in extended logic programming.

Combinations of the above approaches are also proposed in literature, called \emph{hybrid} (\textbf{HB}) configuration systems.
Typically, they use a DL-based representation for the ontology, together with constraints. They combine reasoning engines from these fields to provide inference \cite{HotzKDSN06}.
\subsection{Comparison of approaches}
\citet{Felfernig14} evaluated all these paradigms  with respect to the evaluation criteria from Section \ref{ssec:evaluationcriteria}.
In Table \ref{tab:comparison}, we show this evaluation, together with scores for our implementation in the $\textbf{IDP}$ column, based on the discussion of Section \ref{ssec:evaluation}.

\begin{table}[h]
\centering
\caption{Comparison of systems from Section \ref{sec:related} using criteria from Section \ref{sec:evaluation} as in (Felfernig et al. 2014).
We use a \Checkmark to mark good support, a $\approx$ for partial support and a $-$ to denote that no support is available.}
\label{tab:comparison}
\begin{tabular}{@{}cccccccccccc@{}}
& RBS     & SCSP       & DCSP       & GCSP       & SAT        & FM         & UML        & DL         & ASP        & HB         & IDP        \\ 
C1 & - & -          & -          & -          & -          & \Checkmark & \Checkmark & $\approx$    & -          & $\approx$    &     -       \\ 
C2                      & -                      & -          & -          & \Checkmark & -          & -          & \Checkmark & \Checkmark & \Checkmark & \Checkmark &    \Checkmark        \\
C3                      & -                      & $\approx$    & $\approx$    & $\approx$    & $\approx$    & -          & -          & $\approx$    & $\approx$    & $\approx$    &      $\approx$        \\
C4                      & $\approx$                & -          & -          & \Checkmark & -          & -          & \Checkmark & \Checkmark & \Checkmark & \Checkmark &      \Checkmark      \\
C5                      & -                      & $\approx$    & $\approx$    & $\approx$    & $\approx$    & $\approx$    & $\approx$    & $\approx$    & $\approx$    & $\approx$    &     \Checkmark       \\
C6                      & -                      & \Checkmark & \Checkmark & \Checkmark & \Checkmark & \Checkmark & \Checkmark & \Checkmark & \Checkmark & \Checkmark & \Checkmark \\
C7                      & \Checkmark             & \Checkmark & \Checkmark & \Checkmark & \Checkmark & -          & -          & $\approx$    & $\approx$    & $\approx$    &         $\approx$     \\
C8                      & $\approx$                & -          & -          & \Checkmark & -          & -          & -          & -          & $\approx$    & \Checkmark &      \Checkmark      \\
C9                      & $\approx$                & \Checkmark & \Checkmark & \Checkmark & $\approx$    & -          & -          & \Checkmark & \Checkmark & \Checkmark &     \Checkmark      
\end{tabular}
\end{table}

All these approaches are focused towards one specific inference: ontologies are focused on deduction, rule systems are focused on backward/forward chaining, etc.
These approaches are less general then the KB paradigm, which is specifically designed to reuse the knowledge for different reasoning tasks. 
The contributions of this paper are different from previously discussed approaches: we analyzed  IC problems from a Knowledge Representation point of view.
This paper is a discussion of possible approaches and the importance of this point of view. 
We made a study of desired functionalities for an IC system and how we can define logical reasoning tasks to supply these functionalities.
As far as we are aware, the language we used in this experiment is more expressive than earlier approaches.

The expressivity of the language is crucial for the usability of the approach. 
It allows us to address a broader range of applications, moreover it is easier to formalize and maintain the domain knowledge.
Not discussed by \citet{Felfernig14} et al is work by \citet{ppdp/VlaeminckVD09}. They did a preliminary experiment using the KB approach for interactive configuration, also using the $\fodot$ \idp project. 
It is on this work that we continue in this paper by analyzing a real-life application of a larger scale and discussing new functionalities and inferences.
This theoretical approach benefits from (1) the expressive language to express domain knowledge adequately and (2) the general basic inferences that realise derived inferences in an easy way, supporting the discussed functionalities, resulting in a IC system that scores very well with relation to the evaluation criteria (Table \ref{tab:comparison}).

An interesting remark in Table \ref{tab:comparison} is that the IDP column resembles the GCSP column, a generalisation of CSP, developed for configuration. 
The IDP-system has better support for C5 (maintainability), due to the high level modeling language and the strict seperation between domain knowledge and reasoning. 
GCSP has better efficiency results. This can be partly explained by the fact that CSP uses dedicated algorithms for reasoning over global constraints such as \textit{alldifferent}. 
The goal of reusing knowledge makes that we typically do not make use of this kind of specific algorithms, since a dedicated algorithm can only be developed with one specific inference in mind.
 
\section{Challenges and Future Work} \label{sec:challenge}
Interactive configuration problems are part of a broader kind of problems, namely service provisioning problems. 
Service provisioning is the problem domain of coupling service providers with end users, starting from the request until the delivery of the service. 
Traditionally, such problems start with designing a configuration system that allows users to communicate their wishes, for which we provided a knowledge-based solution.
Once all the information is gathered from a user, it is still necessary to make a plan for the production and delivery of the selected configuration.
Hence the configuration problem is followed by a planning problem
that shares domain knowledge with the configuration problem but that
also has its own domain knowledge about providers of components,
production processes, etc.
This planning problem then leads to a monitoring problem.
Authorizations could be required, payments need to be checked, or it could be that the configuration becomes invalid mid-process.
In this case the configuration needs to be redone, but preferably without losing much of the work that is already done.
Companies need software that can manage and monitor the whole chain, from
initial configuration to final delivery and this without duplication
of domain knowledge. This is a problem area where the KB approach
holds great promise but where further research is needed to integrate
the KB system with the environment that the company uses to follow up its
processes.

Other future work may include language extensions to better support con\-figuration-like tasks.
A prime example of this are templates~\cite{tplp/DassevilleHJD15}. Oftentimes the theory of a configuration problem contains lots of constraints which are similar in structure. It seems natural to introduce a language construct to abstract away the common parts. 
Another useful language extension is reification, to talk about the symbols in a specification rather than about their interpretation.
Reification allows the system to reason on a meta level about the symbol and for example assign symbols to a category like ``Technical'' or ``Administrative''.

 \vspace{-1em}
\section{Conclusion}
\vspace{-0.5em}
The KB paradigm, in which a strict separation between knowledge and problem solving is proposed, was analyzed in a class of knowledge intensive problems: interactive configuration problems. 
As we discussed why solutions for this class are hard to develop, we proposed a novel approach to the configuration problem based on an existing KB system.
We analyzed the functional requirements of an IC system and investigated how we can provide these, using logical inferences on a knowledge base.
We identified interesting new inference methods and applied them to the interactive configuration domain.
We studied this approach in context of a large application, for which we built a proof of concept, using the KB system as an engine, which we extended with the new inferences. 
As proof of concept, we solved a configuration problem for a large banking company.
Results are convincing and open perspectives for further research in service provisioning.

 \bibliographystyle{acmtrans}
\bibliography{krrlib}

\end{document}